\documentclass[runningheads]{llncs}
\usepackage{graphicx}

\usepackage{tikz}
\usepackage{comment}
\usepackage{amsmath,amssymb} 
\usepackage{color}
\usepackage{bbding} 
\usepackage{subcaption} 
\usepackage{array}
\usepackage{float} 

\usepackage[misc]{ifsym}

\usepackage{amsmath,amsfonts,bm}

\def\eqref#1{equation~\ref{#1}}

\def\1{\bm{1}}

\newcommand{\test}{\mathcal{D_{\mathrm{test}}}}

\def\vb{{\bm{b}}}

\def\mA{{\bm{A}}}

\def\mW{{\bm{W}}}

\DeclareMathAlphabet{\mathsfit}{\encodingdefault}{\sfdefault}{m}{sl}
\SetMathAlphabet{\mathsfit}{bold}{\encodingdefault}{\sfdefault}{bx}{n}

\def\sR{{\mathbb{R}}}

\usepackage[accsupp]{axessibility}  

\begin{document}
\pagestyle{headings}
\mainmatter
\def\ECCVSubNumber{5704}  

\title{Towards Accurate Binary Neural Networks via Modeling Contextual Dependencies} 


\titlerunning{Towards Accurate BNNs via Modeling Contextual Dependencies}
%
\author{Xingrun Xing\inst{1} \and
Yangguang Li\inst{2} \and
Wei Li\inst{3} \and
Wenrui Ding\inst{1} \and
Yalong Jiang\thanks{Corresponding author.}\inst{1} \and
Yufeng Wang\inst{1} \and
Jing Shao\inst{2} \and
Chunlei Liu\inst{1} \and
Xianglong Liu\inst{1}}
\authorrunning{X. Xing et al.}
%
\institute{Beihang University\\
\email{\{sy2002215, ding, allenyljiang, wyfeng, liuchunlei, xlliu\}@buaa.edu.cn} \and
SenseTime Group\\
\email{liyangguang@sensetime.com, shaojing@senseauto.com} \and
Nanyang Technological University\\
\email{wei.l@ntu.edu.sg}}
\maketitle


\begin{abstract}
Existing Binary Neural Networks (BNNs) mainly operate on local convolutions with binarization function. However, such simple bit operations lack the ability of modeling contextual dependencies, which is critical for learning discriminative deep representations in vision models. 
In this work, we tackle this issue by presenting new designs of binary neural modules, which enables BNNs to learn effective contextual dependencies. 
First, we propose a binary multi-layer perceptron (MLP) block as an alternative to binary convolution blocks to directly model contextual dependencies. Both short-range and long-range feature dependencies are modeled by binary MLPs, where the former provides local inductive bias and the latter breaks limited receptive field in binary convolutions. 
Second, to improve the robustness of binary models with contextual dependencies, we compute the contextual dynamic embeddings to determine the binarization thresholds in general binary convolutional blocks.
Armed with our binary MLP blocks and improved binary convolution, we build the BNNs with explicit Contextual Dependency modeling, termed as BCDNet. 
On the standard ImageNet-1K classification benchmark, the BCDNet achieves 72.3\% Top-1 accuracy and outperforms leading binary methods by a large margin. In particular, the proposed BCDNet exceeds the state-of-the-art ReActNet-A by 2.9\% Top-1 accuracy with similar operations. Our code is available at \url{https://github.com/Sense-GVT/BCDNet}.
\end{abstract}

\section{Introduction}

Over the last decade, deep learning methods have shown impressive results for a multitude of computer vision tasks. However, these models  require massive parameters and computation to achieve strong performance, hindering their application in practical embedded devices with limited storage and computing resources.
Aiming at reduce computational cost in deep models, model compression has drawn growing interests and developed into various methods, such as quantization \cite{liu2021post,lee2021network,zhao2021distribution}, pruning \cite{joo2021linearly,yamamoto2021learnable}, knowledge distillation \cite{ji2021refine,song2021robust}, etc. 
Binary neural networks (BNNs) \cite{liu2020reactnet,jiang2021training} can be viewed as an extreme case of low-bit quantization and become one of the most prevailing approaches to save computational cost at an extremely high rate. 
%
BNNs binarize the weights and activations to save at most 64$\times$ operations and 32$\times$ memory. However, current BNNs for vision tasks are built upon simple local convolutions, contributing to limited perception field in each layer. As such, it is of significant interest to explore more efficient contextual binary operations for BNNs.

\begin{figure}[t]

    \begin{center}

    \includegraphics[width=0.7\linewidth]{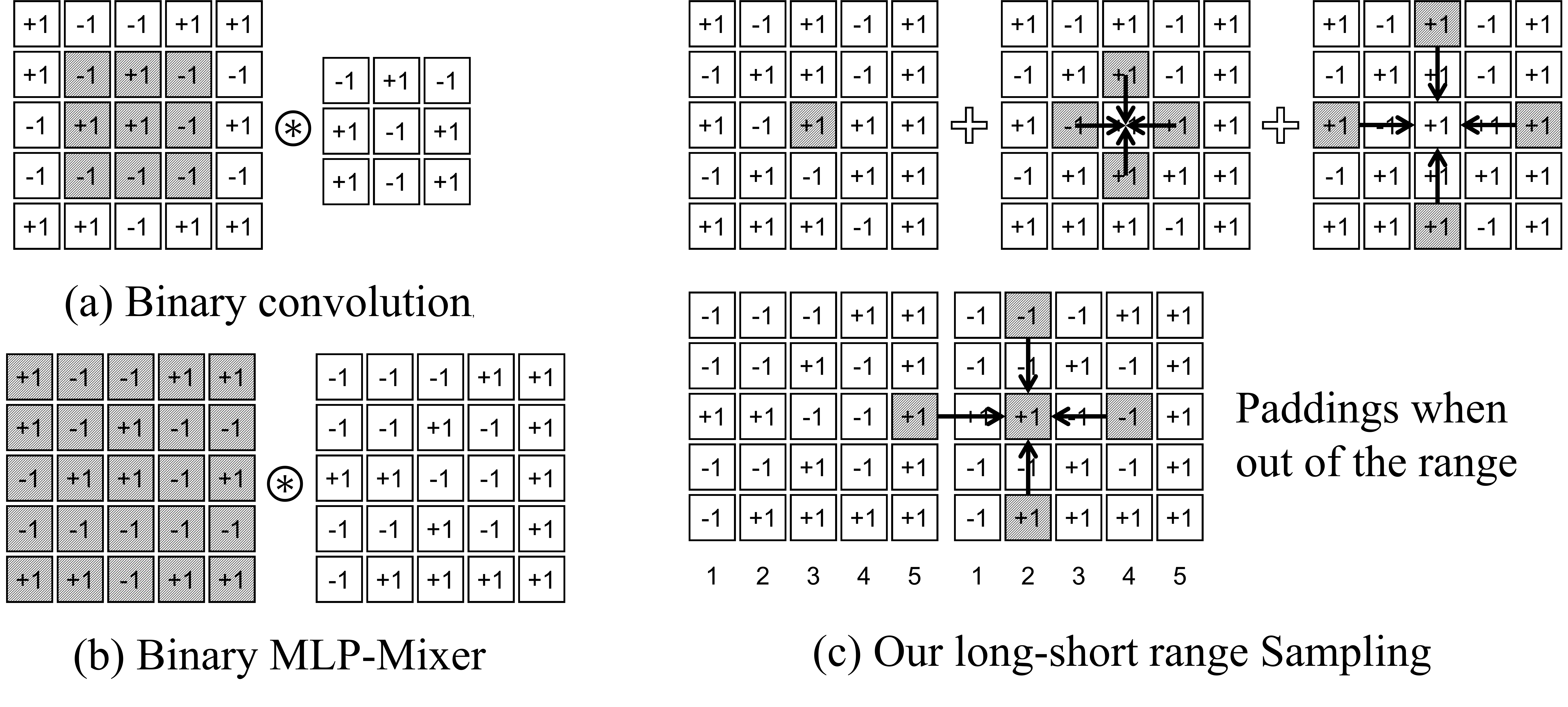}
    \end{center}
    \caption{Comparisons of binary operations: (a) The binary convolutions have inductive bias but limited local perceptions; (b) The binary token-mixing MLPs are sharing contextual perceptions but difficult to optimize; (c) Our proposed binary MLPs achieve inductive bias in short-range and explore long-range dependencies concurrently.}
\label{fig:motivation}
\end{figure}

Despite the advantage of inductive bias (e.g. locality) of convolutions for vision tasks, recent studies have shown improving interests in contextual operations, including the vision transformers (ViTs) and MLPs, that achieves state-of-the-art performance. 
Different from local convolutions, ViTs and MLPs treat images as tokens and model the dependencies across all tokens at once.
%
To explore the binary contextual interaction in BNNs, we start by implementing a vanilla binary MLP-Mixer model, which we find is difficult to optimize compared with traditional binary convolutions. The model also fails to improve BNN accuracy with the help of contextual dependencies. 
%
This suggests local inductive bias is more crucial for BNNs compared with real-valued models.

Motivated by this, we aim to take advantage of contextual information while maintaining local inductive bias concurrently. A long-short range binary MLP operation is first proposed as an alternative to binary convolutions, where the short-range and long-range dependencies are introduced upon the standard binary convolutions, as shown in Fig. \ref{fig:motivation}(c). We indicate the original location modeling as pointwise branch. The pointwise and short-range dependencies provide inductive bias similar to convolutions, meanwhile we break the local perceptions with long-range modeling. 
In summary, the three modeling patterns jointly perform a contextual interaction at each location.
In this way, the proposed long-short range binary MLPs take the essence of convolutions and MLP-Mixer while alleviate their problems (Fig. \ref{fig:motivation} (a) and (b)).

\begin{figure*}[t]

    \begin{center}
    \includegraphics[width=0.7\textwidth]{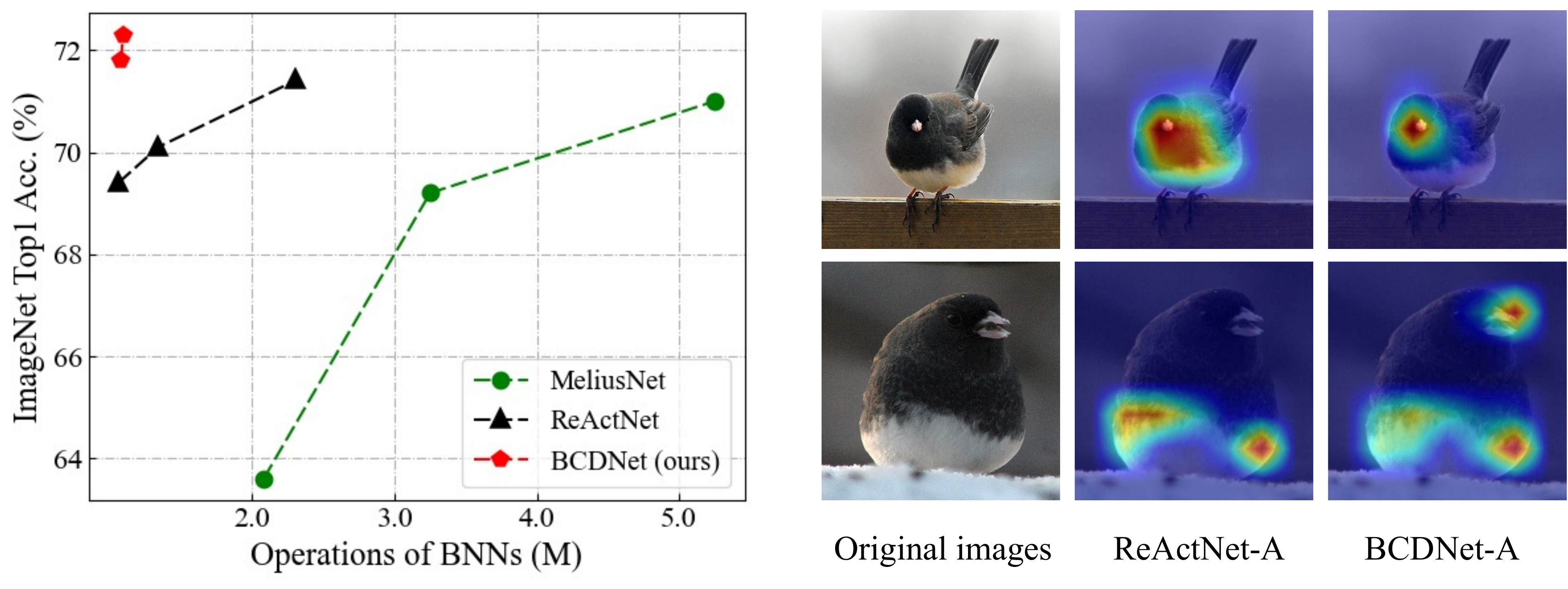}
    \end{center}
    \caption{The performance of binary sparse MLPs express state-of-the-art binary CNNs; class activation maps (CAM) suggest that contextual binary MLPs are better to detect salient features, for example the 'beak' for the 'bird' class.}
    \label{fig:perf_acts}
\end{figure*}

Furthermore, we leverage the proposed long-short-range binary MLP blocks to replace some of the last convolutional blocks to improve the contextual perception field of BNNs while maintaining almost the same calculation.
%
In addition, to access global perceptions in general binary convolutions, we further propose to compute the contextual dynamic embeddings to determine binarization thresholds. In binary convolutional blocks, we first aggregate global features with global average pooling, and then transform pooled features as global embeddings, which adapts binarization at inference.
%
With the binary contextual MLPs and dynamic convolutions, we present a new binary networks named BCDNet to enhance the \textbf{B}inary \textbf{C}ontextual \textbf{D}ependencies that is deficient in previous binary CNNs. 
As shown in Fig. \ref{fig:perf_acts}, with similar architecture and computational cost, BCDNets achieves 2.9\% accuracy gain against the state-of-the-art ReActNet-A model \cite{liu2020reactnet}. It also detects more discriminative features in the contextual range.

Our contributions are summarized as follows:

$\bullet$ We make the first attempt to break local binary convolutions and explore more efficient contextual binary operations.

$\bullet$ We design a long-short range binary MLP module to enhance modeling contextual dependencies in BNNs.

$\bullet$ To our best knowledge, the proposed BCDNet reports the best BNN performance on the large-scale ImageNet dataset currently.


\section{Related Work}

\subsection{Binary Neural Networks}
1-bit neural networks are first introduced by BNN \cite{courbariaux2016binarized} but encounters severe quantity error, especially in large scale datasets. To improve performance, some works focus to optimize binary weights. For example, XNOR-Net \cite{rastegari2016xnor} reduce binarization error with scaling factors; RBNN \cite{lin2020rotated} learns rotation matrixes to reduce angular bias. SA-BNN \cite{liu2021sa} reduces flips in training; AdamBNN \cite{liu2021how} explores influence of optimizers. Others try to modify architectures to improve expression ability. For instance, Bi-Real Net \cite{liu2018bi} adds more skip connections and achieves better performance; ReActNet \cite{liu2020reactnet} employs channel shift and the MobileNet-V1 \cite{howard2017mobilenets} architecture; CP-NAS \cite{li2020cp} adopts architecture search to explore connections. However, previous works only focus on CNN-based BNNs and ignore the performance upper-bound of binary convolutions. This work provides MLP based BNNs and exceed state of the arts.

\subsection{Vision Transformers and MLPs}

Recently, transformers \cite{zheng2021rethinking,lin2021end,dai2021up} are widely used to explore contextual dependencies. ViTs \cite{dosovitskiy2020image} first define to represent an image as 16x16 tokens using an embedding layer, and then stack self-attention and MLP layers. To improve performance of ViTs, DeiTs \cite{deit} propose a distillation token which benefits transformers a lot. More recently, Swin transformers \cite{liu2021Swin}, Twins \cite{chu2021twins} and NesT \cite{zhang2021aggregating} explore various self-attentions to save computational cost, which makes transformers become the first choice in vision tasks. At the same time, works such as MLP-Mixers \cite{tolstikhin2021mlp}, ResMLP \cite{ResMLPFN} simplify transformers using spatial and channel MLPs, but encounter overfeating. To this end, works such as $S^2$MLPs \cite{yu2021s}, CycleMLPs \cite{chen2021cyclemlp} adopt surrounding shift operations to replace token-mixing MLPs. However, these works only focus on shifts and full precision circumstances. For BNNs, there still lacks in efficient contextual operations.


\section{Background}

In this section, we first review the binary convolution in recent binary convolutional neural networks \cite{courbariaux2016binarized,rastegari2016xnor,liu2020reactnet}. Then, we follow the design in the binary convolution and make an initial attempt to binarize the MLP layer in vision MLP models \cite{tolstikhin2021mlp,ResMLPFN,yu2021s,chen2021cyclemlp}. We further conduct pilot studies to evaluate the performance of a binarized MLP-Mixer \cite{tolstikhin2021mlp} and a ResNet$+$MLP architecture on ImageNet.

\subsection{Binary Operations}

\noindent\textbf{Binary Convolution.} 
By replacing arithmetic operations with more efficient bitwise operations, binary convolutions can drastically reduce memory allocation and accesses.
Typically, full precision weights $\mW$ and activations $\mA$ are first binarized as $\mW_b$ and $\mA_b$ using the quantization function $Q_b(\cdot)$. As such, convolutions can be implemented as XNOR-Bitcounting operations \cite{courbariaux2016binarized,rastegari2016xnor}:
\begin{equation}  \label{eq_conv}
\texttt{BiConv}(\mA)=\alpha {Q_b(\mW)}^TQ_b(\mA)=\frac{{\left\lVert \mW \right\rVert}_{\ell_1}}{k\times k\times c_{in}} \texttt{bitcount}({\mW_b}^T\oplus \mA_b),
\end{equation}
where $k$ and $c_{in}$ indicate the kernel size and input channels respectively, and $\alpha $ serves as the scale factor $\frac{{\left\lVert \mW \right\rVert}_{\ell_1}}{k\times k\times c_{in}}$. 
Following \cite{liu2018bi}, the quantization function $Q_b(\cdot)$ can be implemented in the form of a differentiable polynomial function:
\begin{align}
\label{eq_dq}
Q_b(x) = \left\{  
                \begin{array}{lr}  
                - 1 & {\rm if} \ x  < -1 \\ 
                2x+x^2 \ \ &{\rm if} -1 \leqslant x < 0 \\
                2x-x^2 &{\rm if} \ 0 \leqslant x < 1 \\
                1 & {\rm otherwise}  
                \end{array} 
\right. 
,
\quad 
\frac{\partial Q_b(x)}{\partial x} = \left\{  
                \begin{array}{lr}  
                2+2x \ \ &{\rm if} -1 \leqslant x < 0 \\
                2-2x &{\rm if} \ 0 \leqslant x < 1 \\
                0 & {\rm otherwise}  
                \end{array} 
\right. 
,
\end{align}
where gradients are approximated during back-propagation to optimize model weights $\mW$.
Compared with real-valued convolutions with the 32-bit weight parameters, binary convolutions output $1$-bit representations and can obtain up to 32$\times$ memory and 64$\times$ operation savings, leading to much faster test-time inference and lower power consumption.
    
\begin{figure}[t]
    \begin{center}
    \includegraphics[width=0.8\linewidth]{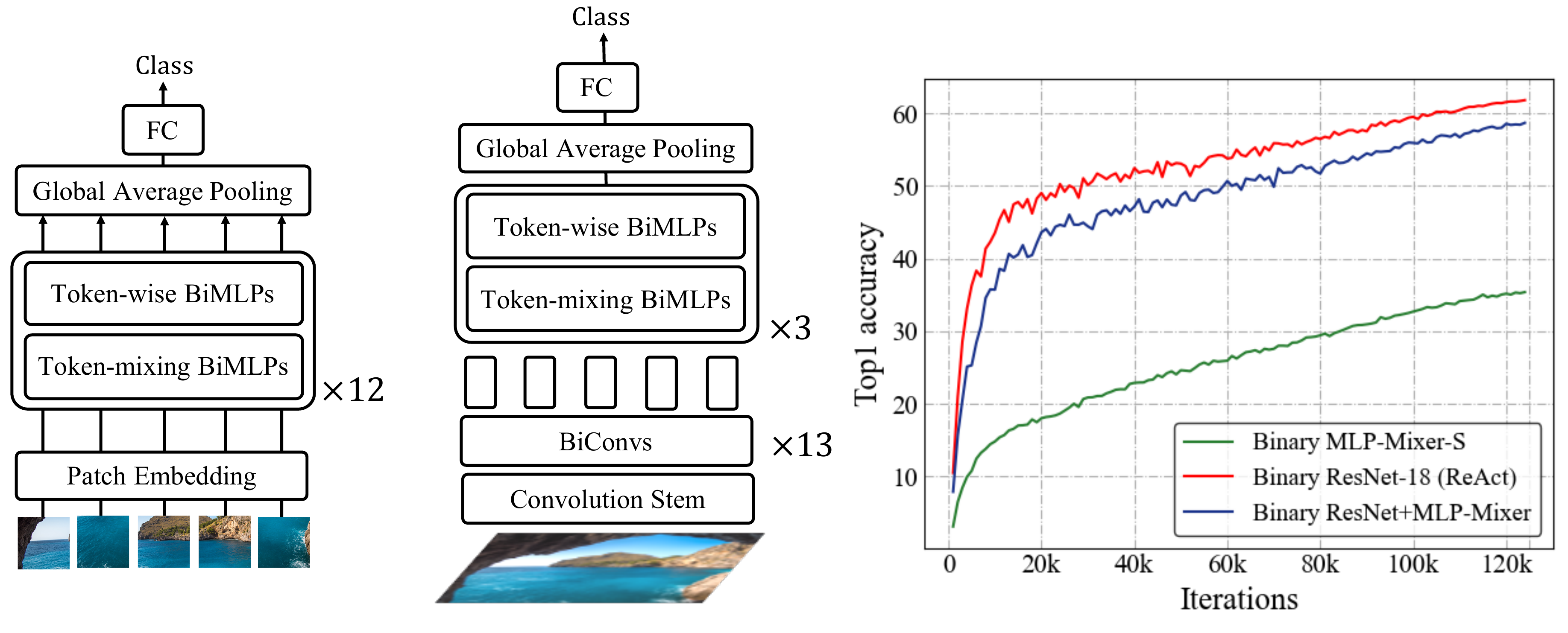} 
    \end{center}
    \caption{Initial implementations of binary MLP-Mixer (left) and a binary ResNet adding MLP architecture (middle). The comparison of performance with binary MLP-Mixer-S, ResNet-18 and ResNet$+$MLP is shown on the right.}
    \label{fig:pilot_study}
\end{figure}

\noindent\textbf{Binary MLP.}
The MLP layer conducts regular matrix multiplications, which can be treated as $1\times1$ convolution that are repeatedly applied across either spatial locations or feature channels. For spatial dimensions, A single token-mixing MLP layer can be binarized in the form of:  
\begin{equation}  \label{eq_mlp}
\texttt{TM-BiMLP}(\mA)=\frac{{\left\lVert \mW \right\rVert}_{\ell_1}}{h\times w} \texttt{bitcount}({\mW_b}^T\oplus \mA_b),
\end{equation}
where $h$ and $w$ indicate the height and width of input features. Next, we binarize MLP-Mixer-S as shown in Fig. \ref{fig:pilot_study} (left). For binary MLP-Mixer, one MLP block is formulated by 2 binarized token-mixing MLP layers and 2 binarized token-wise MLP layers, and the expending ratio is 4. Later, we further explore a unified architecture composed of binary convolutions and MLPs. As shown in Fig. \ref{fig:pilot_study} (middle), our initial attempt is to replace the last 3 binary convolution blocks as binarized MLP-Mixer blocks according to similar operations, which aims at modeling contextual dependencies in high level. Because MLP-Mixer blocks have large expending ratio and multiple feed-forward layers, the binarized MLP-Mixer block presents similar or larger model size than the binary convolution.

\subsection{Analysis}
In order to evaluate the performance of token-mixing binary MLP layers introduced above, we binarize operations in MLP-Mixer-S \cite{tolstikhin2021mlp} except the first and last layers. For comparison, we also evaluate binary ResNet-18 implemented by ReActNet and a binary ResNet$+$MLP architecture mentioned above, where the convolutional and MLP blocks are implemented by the ReActNet and binarized MLP-Mixer ones respectively.
We train binary models on the public ILSVRC2021 ImageNet dataset \cite{russakovsky2015imagenet} for 100 epochs. Complete training configurations appear in the Supplementary A.

\noindent\textbf{Summary.} 
As shown in Fig. \ref{fig:pilot_study}, although MLP-Mixers achieves larger perception field than CNNs, the binary version cannot even achieve 40\% top1 accuracy, which is worse than the first BNN model~\cite{courbariaux2016binarized}. Compared with the binary MLPs that directly omit inductive bias, binary convolutions achieve best accuracy with the help of local perceptions and transformation invariance. Moreover, in the unified architecture of binary convolutions and MLPs, binary token-mixing layers still can not benefit from contextual perceptions, but much easier to train than stand-alone binary MLPs. Here we draw the following analysis:

$\bullet$ In the design of contextual binary operations, one need consider the inductive bias sufficiently. However, only preventing local dependencies in binary convolutions also barriers BNNs to capture long-range interactions. To this end, comprehensive design of local and contextual modeling is desired.

$\bullet$ In the design of BNN architectures, binary convolution stage is necessary, as stand-alone binary MLP-Mixers are difficult to optimize.

\section{Method}

\begin{figure*}[t]
    \centering{
    \centerline{\includegraphics[width=1\textwidth]{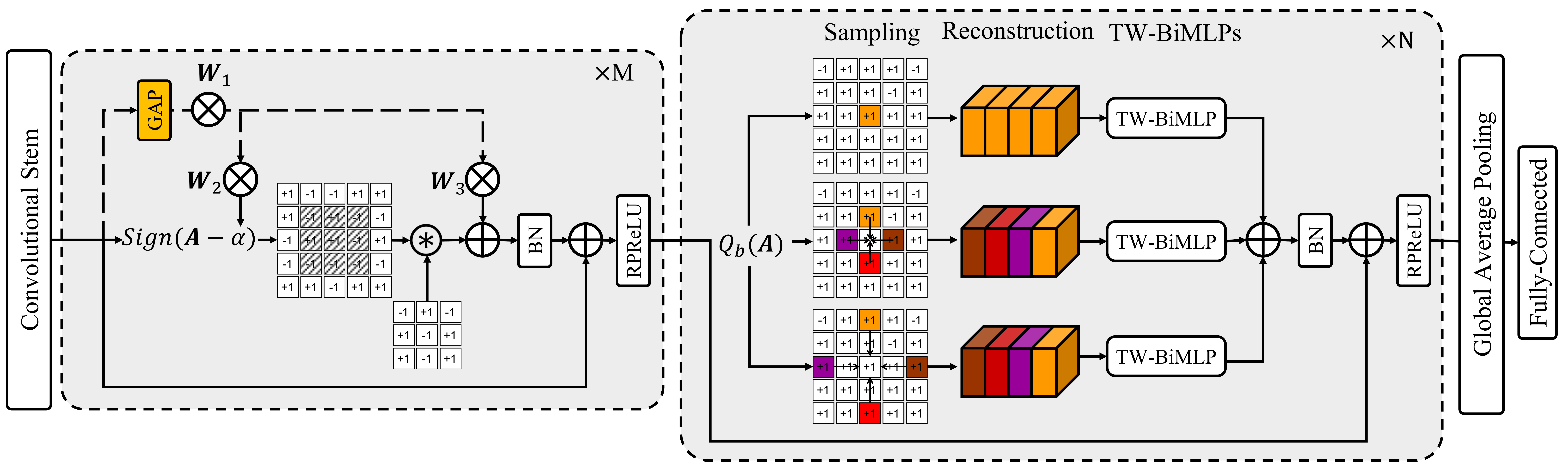}}
    \caption{Overall architecture of BCDNet, which consists of $M$ binary convolution blocks and $N$ binary MLP blocks. In practice, $M=11$ and $N=9$.}
    \label{fig:overall_block}
    }
\end{figure*}

Our proposed binary model is built on modeling contextual dependencies. An overview of the BCDNet architecture is illustrate in Fig. \ref{fig:overall_block}. We first describe the fundamentals of binary MLP block. Then, we introduce the contextual design of binary convolutions with dynamic embeddings. Finally, we present the complete BCDNet architecture with modeling contextual dependencies.

\subsection{Binary MLP Block with Long-Short Range Dependencies}

We first introduce three basic components in our binary MLP blocks, including a long-short range sampling scheme, a token reconstruction function and binary MLP operations, and then indicate the overall binary MLP block.

\subsubsection{Long-Short Range Sampling.}
As shown in Fig. \ref{fig:motivation}(c), we design a comprehensive sampling function which consists of three sampling patterns: the pointwise, short-range and long-range samplings. First, pointwise and short-range tokens provide enough local inductive bias like convolutions. Second, long-range tokens are responding to explore contextual perceptions like self-attentions and token-mixing MLPs. In every block, one pointwise, four short-range, and four long-range tokens are sampled at each location, which compose a sparse contextual sampling. Given a sampling distance $(r_1,r_2)$ in spatial, we use a sampling function to indicate the index of a token shifted from a location $(x_1,x_2)$:
\begin{equation} \label{eq4}
S\left(r_{1}, r_{2}\right)=\left\{y: y=\left(x_{1}+r_{1}, x_{2}+r_{2}\right)\right\}.
\end{equation}
In each location, we donate sampled long-range, short-range and the pointwise tokens as $\{S(0, \pm h/2), S(\pm w/2, 0)\}$, $\{S(0, \pm 1), S(\pm 1, 0)\}$ and $S(0, 0)$ respectively. Note that, when an index comes out of range, we just pad features from the other side and regard the index as recurrent (shown in Fig. \ref{fig:motivation} (c)).

\subsubsection{Token Reconstruction across Samplings.}
Based on sampled tokens above, one simple way to aggregate them is concatenation. However, this will increase the channel numbers and calculations. To save calculations, we select and concatenate parts of channels of sampled tokens similar to \cite{wu2018shift}, which keeps channels unchanged in reconstructed tokens. We choose the spatial shift operation \cite{wu2018shift} to reconstruct four sampled long-range tokens or four short-range tokens respectively, which is zero parameters and FLOPs. In each location, we can reconstruct a new short-range token $\mA_b^S$ from parts of four sampled short-range tokens:
\begin{eqnarray} \label{eq5}
    \mA_{b}^{S}=&Cat\{\mA_{b}[0: c / 4]_{S(-1,0)},\mA_{b}[c / 4: c / 2]_{S(1,0)},\nonumber \\
&\mA_{b}[c / 2: 3 c / 4]_{S(0,-1)},\mA_{b}[3 c / 4: c]_{S(0,1)}\},
\end{eqnarray}
where $\mA_b$ is a binarized input token, $S(r_1, r_2)$ indexes its location, and $[:]$ indexes its channels. Similarly, a new long-range token $\mA_b^L$ is reconstructed from parts of four long-range tokens:
\begin{eqnarray} \label{eq6}
    \mA_{b}^{L}=&Cat\{\mA_{b}[0: c / 4]_{S(-h / 2,0)}, \mA_{b}[c / 4: c / 2]_{S(h / 2,0)}, \nonumber \\
&\mA_{b}[c / 2: 3 c / 4]_{S(0,-w / 2)}, \mA_{b}[3 c / 4, c]_{S(0, w / 2)}\}.
\end{eqnarray}
Here, we set the short-range interaction distance as one token and long-range as h/2 or w/2 tokens in space. In summary, we can obtain an pointwise token, a reconstructed short-range token and a long-range token $(\mA_b, \mA_b^S, \mA_b^L)$ for each location, so that a sparse contextual token sampling is formulated.


\subsubsection{Token-wise Binary MLP.}

In the following, we model three kinds of sampled tokens $(\mA_b, \mA_b^S, \mA_b^L)$ independently using three token-wise binary MLPs. In MLP blocks, We follow ReActNets \cite{liu2020reactnet} and apply $Sign(.)$ and $RSign(.)$ to binarize weights and activations respectively. Given binarized weights $\mW_b$ and activations $\mA_b$, the token-wise binary MLP is defined as a binarized fully connected layer across the channel dimension:
\begin{equation}  \label{eq2}
\texttt{TW-BiMLP}(\mA)=\frac{{\left \| \mW \right \|}_{\ell_1}}{c_{in}} \texttt{popcount}({\mW_b}^T\oplus \mA_b),
\end{equation}
where ${\left \| \mW \right \|_{\ell_1}}$ indicates the L1 norm of real valued weights across the channel dimension. Similar to XNOR-Net \cite{rastegari2016xnor}, we compute a scaling factor $\frac{{\left \| \mW_{b} \right \|}_{\ell_1}}{c_{in}}$ to minimize binarization error in weights. Next, the binary matrix product can be implemented by XNOR and bit-counting. The difference between Eq. \ref{eq_mlp} and \ref{eq2} is fully connected dimensions.


\subsubsection{Binary MLP Block.}
To explore block-wise contextual interactions, we propose the long-short range binary MLP block as an alternative to binary convolutional residual blocks in a network. Different from real-valued MLP-Mixers that aggregate all tokens indistinctively, our primary design is independently sampling pointwise, short-range, and long-range tokens according to a long-short sparse sampling rule. After a token reconstruction function to downsample channels, we use three token-wise binary MLPs to model three kinds of dependencies comprehensively. As shown in Fig. \ref{fig:overall_block}, we first binarize activations using RSign(·). In every location, three binary MLPs $\{P, S, L\}$ independently model three kinds of dependencies and are then added together:
\begin{equation} \label{eq7}
    \mA^{i}=P\left(\mA_{b}^{i-1}\right) + S\left(\mA_{b}^{S}\right) + L\left(\mA_{b}^{L}\right),
\end{equation}
where ${\{\mA_b,\mA_b^S,\mA_b^L\}}$ indicates pointwise binary tokens, reconstructed short-range tokens and long-range tokens respectively. Like residual blocks, batch norm, skip connection and activation functions are also attached as in Fig. \ref{fig:overall_block}.

When the channel size is large (e.g., 512 and 1024), a binary MLP block consumes about 1/3 operations compared with a typical $3\times3$ binary convolution with the same input and output channels. Our binary MLPs can be easily adopted as stand-alone or plug-in building blocks in BNNs With similar configurations of regular layers.

\subsection{Binary Convolutions with Dynamic Contextual Embeddings}

To enhance model robustness with contextual dependencies, we improve the binary convolutions by introducing dynamic embedding to determine binarization thresholds $ \alpha \in \sR^{c}$ before the sign functions. Unlike ReActNets \cite{liu2020reactnet} that learns overall thresholds from the whole dataset without representing image-specific characteristics, we propose to infer binarization thresholds according to the input images dynamically. Specifically, in each residual block, we first obtain the mean value of each channel using the global average pooling ($\texttt{GAP}$), followed by the matrix product operation:
\begin{equation} \label{eq10}
\alpha(\mA) = \texttt{GAP}(\mA) \mW_{1}+ \vb_{\alpha},
\end{equation}
where $\mA \in \sR^{c \times h\times w}$ is an input tensor, and $\{\mW_1\in \sR^{c\times \frac{c}{4}}, \vb_\alpha \in \sR^{\frac{c}{4}}\}$ are learnable parameters. Note that, this matrix product increases negligible calculation due to the GAP function. Based on the contextual feature $\alpha(\mA)$, the channel-wise thresholds $\beta(\mA)\in \sR^{c}$ is obtained by a matrix product operation:
\begin{equation} 
\label{eq11}
\beta(\mA)=\alpha(\mA) \mW_{2} + \vb_{\beta},
\end{equation}
with learnable parameters ${\{\mW_2 \in {\sR^{\frac{c}{4} \times c} }, {\vb_\beta} \in \sR^c\}}$. After merging $\beta(\mA)$ with the following sign function, we quantize inputs using instance specific thresholds:
\begin{eqnarray} 
\label{eq12}
Q_{b}(x) = \texttt{sign}(x-\beta(\mA))=\left\{\begin{array}{c}
+1, x>\beta(\mA) \\
-1, x \leq  \beta(\mA).
\end{array}\right.
\end{eqnarray}

To compensate the activation distribution variance caused by subtracting binarization thresholds, we learn additional embedding $\gamma(\mA)$  after binary convolutions:
\begin{equation}
\gamma(\mA)=\alpha(\mA) \mW_{3} + \vb_{\gamma},
\end{equation}
where dynamic embeddings ${\gamma(\mA) \in \sR^{c^{\prime}}}$ are also transformed by a matrix product with parameters ${\mW_3 \in \sR^{{\frac{c}{4}}\times c^{\prime}}}$, ${{\vb_\gamma }\in \sR^{c^{\prime}}}$. 
As such, the output tensor $\mA^{\prime}$ is given by:
\begin{equation}
\mA^{\prime}=\mA^{\prime}+\gamma(\mA),
\end{equation}
In summary, two dynamic embeddings that model contextual dependencies are attached before and after the binary convolution respectively, formulating a dynamic binarization strategy for given inputs.

\subsection{Overall Network}

In Fig. \ref{fig:overall_block}, we outline the overall architecture of BCDNet. BCDNet is composed of two stages: the binary CNN-embedding stage with M binary convolutional blocks, and the binary MLP stage with N proposed binary MLP blocks. In a standard ViT or MLP architecture, the first embedding layer are implemented by a non-overlapping stride convolution. However, as suggested by \cite{xiao2021early}, this patchify stem layer leads to optimization problems and can be sensitive to training strategies. In section 3.2, we find stand-alone contextual binary models are difficult to optimize. Instead, we adopt our binary MLP modules to enhance high level features. First, we replace several binary convolution blocks with our binary MLPs based on ReActNet-A, which we indicate as BCDNet-A. To exploit the trade-off between convolution and MLP blocks, we perform a grid search under the same computational cost, and determine 11 convolutional blocks and 9 MLP blocks in Fig. \ref{fig:fig5}. Second, we apply both improved binary convolution and MLP blocks and introduce a BCDNet-B model to further improve performance.

\section{Experiments}

\subsection{Implementation Details}
We train and evaluate the proposed BNNs on the large-scale ImageNet-1K dataset \cite{russakovsky2015imagenet}, which contains 1.28M images for training and 50K images for validation. 
We use $224 \times 224$ resolution images with standard data augmentations similar to Real-To-Binary \cite{martinez2020training}. Following ReActNet~\cite{liu2020reactnet} and Real-to-Binary \cite{martinez2020training}, we conduct a two-step training strategy with knowledge distillation and weight decay. We first train binary activation and real-valed weight networks for 300K iterations. Then, we initialize model weights from the first step, and train binary weight and activation models for 600K iterations. We use the AdamW \cite{loshchilov2017decoupled} optimizer with a cosine learning rate schedule and batch size of 256. In two steps, we set the learning rate ${1\times{10}^{-4}}$ and ${5\times{10}^{-5}}$ and the weight decay $1\times{10}^{-5}$ and $0$, accordingly. For supervision, we use real labels with smooth rate 0.1 and distill from a full precision ResNet50. Detailed evaluation of distillation affect is shown in Supplementary B. For dynamic embeddings, we initialize $W_2, W_3$ and all biases as zeros and finetune another 300K iterations in Table \ref{table1}.

Considering one binary MLP block has almost $1/3$ operations of a $3 \times 3$ binary convolution block, BCDNet-A simply replace the last three $3 \times 3$ convolution blocks in ReActNet-A with 9 binary MLP blocks and keeps almost the same computational cost (Table \ref{table1}); $1 \times 1$ binary convolutions in ReActNet-A remain unchanged. The replacing number is determined by a grid search in Fig. \ref{fig:fig5}. Based on BCDNet-A, we further attach contextual embeddings in the CNN stage to improve binary convolutions and formulate BCDNet-B. We follow previous works \cite{bethge2020meliusnet} and use BOPs, FLOPs and OPs to measure the computational cost, where OPs=1/64BOPs + FLOPs. Note that the FLOPs indicate full precision MACs in BNNs as \cite{liu2018bi}.

\begin{table*}[t]
    \caption{Comparisons with state of the arts. “W/A” is bit-width of weights and activations. We underline ReActNet-A with consideration of our models share the similar architecture and operations. “\dag” indicates operations reported by ReActNets which may omit some small operations. We also report operations of BCDNets only considering conv. and fc. in “(.)”.}
    \label{table1}
    \begin{center}
    \resizebox{\textwidth}{!}{
    \begin{tabular}{lcccccr}
    \hline
    Methods & W/A & BOPs ${(\times{10}^9)}$ & FLOPs ${(\times{10}^8)}$ & OPs ${(\times{10}^8)}$& Top1 (\%) & Top5 (\%) \\
    \hline
    Mobile-V1 \cite{howard2017mobilenets} & 32/32 & 0 & 5.69 & 5.69 & 70.6 & {--} \\
    Mobile-V2 \cite{sandler2018mobilenetv2} & 32/32 & 0 & 3.00 & 3.00 & 72.0 & {--} \\
    ResNet-18 \cite{he2016deep}    & 32/32 & 0 & 18.14 & 18.14 & 69.3 & 89.2 \\
    \hline
    BWN \cite{rastegari2016xnor} & 1/32 &{--} &{--} &{--} &60.8 &83.0 \\
    LQ-Net \cite{zhang2018lq}  &1/2  &{--} &{--} &{--}	 &62.6 &84.3 \\
    DoReFa \cite{zhou2016dorefa}  &2/2 &{--} &{--} &{--}	&62.6 &84.4 \\
    SLB \cite{yang2020searching} &1/8 &{--} &{--} &{--}	&66.2 &86.5 \\
    \hline
    ABC-Net \cite{lin2017towards} & (1/1)$\times 5$ & {--} & {--} & {--} & 65.0 & 85.9\\
    Bi-Real-34 \cite{liu2018bi} & 1/1 & 3.53 & 1.39 & 1.93 & 62.2 & 83.9 \\
    Real-to-Bin \cite{martinez2020training} & 1/1 & 1.68 & 1.56 & 1.83 & 65.4 & 86.2\\
    FDA-BNN* \cite{xu2021learning} & 1/1 & {--} & {--} & {--} & 66.0 & 86.4\\
    SA-BNN-50 \cite{liu2021sa} & 1/1 & {--} & {--} & {--} & 68.7 & 87.4\\
    MeliusNet-22 \cite{bethge2020meliusnet} & 1/1 & 4.62 & 1.35 & 2.08 & 63.6 & 84.7\\
    MeliusNet-42 \cite{bethge2020meliusnet} & 1/1 & 9.69 & 1.74 & 3.25 & 69.2 & 88.3\\
    MeliusNet-59 \cite{bethge2020meliusnet} & 1/1 & 18.3 & 2.45 & 5.25 & 71.0 & 89.7\\
    ReActNet-A \cite{liu2020reactnet} & 1/1 & \underline{4.82} & \underline{0.31 (0.12\dag)} & \underline{1.06 (0.87\dag)} & \underline{69.4} & {--}\\
    ReActNet-B \cite{liu2020reactnet}& 1/1 & {4.69} & {0.61 (0.44\dag)} & {1.34 (1.17\dag)} & {70.1} & {--}\\
    ReActNet-C \cite{liu2020reactnet}& 1/1 & {4.69} & {1.57 (1.40\dag)} & {2.30 (2.14\dag)} & {71.4} & {--}\\
    \hline
    \textbf{BCDNet-A} & \textbf{1/1 }& \underline{\textbf{4.82}} & \underline{\textbf{0.32 (0.12)}} & \underline{\textbf{1.08 (0.87)}} & \underline{\textbf{71.8 (+2.4)}} &\textbf{ 90.3}\\
    
    \textbf{BCDNet-B} & \textbf{1/1} & \underline{\textbf{4.82}} & \underline{\textbf{0.34 (0.14)}} & \underline{\textbf{1.09 (0.89)}} &\underline{\textbf{ 72.3 (+2.9)}} & \textbf{90.5}\\
    \hline
    \end{tabular}
}
\end{center}
\end{table*}

\subsection{Comparisons with State of the Arts}
In Table \ref{table1}, we present comparisons of BCDNets with state-of-the-art binary or low-bit neural networks. To our best of knowledge, all BNNs are based on convolutional networks. BCDNets achieve the highest accuracy, which indicates the necessity to improve binary convolutions with contextual interactions. With similar architecture and computational cost, we observe that BCDNet-A is able to exceed ReActNet-A by a large margin (+2.4\% top1 accuracy). Moreover, with additional dynamic embeddings, we find BCDNet-B further improves 0.5\% top-1 accuracy with a little computational overhead, which validates the effectiveness of contextual dynamic embeddings. Also, BCDNet-B outperforms the full-precision MobileNet-V2 for the first time with a binary model.

\begin{table}[!hbt]
    \caption{Comparison with ResNet-18 architectures on the ImageNet dataset. W/A indicates bit-width of weights and activations respectively.
    }
    \label{Table2}
    \begin{center}
    \setlength{\tabcolsep}{2mm}{
    \begin{tabular}{lcr||lcr}
    \hline
    Methods & W/A & Top1 Acc. & Methods &	W/A &	Top1 Acc. \\
    \hline
    ResNet-18 &	32/32&	69.3&	RBNN \cite{lin2020rotated}&	1/1& 59.9 \\
    XNOR-Net \cite{rastegari2016xnor}&	1/1& 51.2& SA-BNN \cite{liu2021sa}& 1/1& 61.7 \\
    Bi-Real \cite{liu2018bi}&	1/1&	56.4&	ReActNet (R18) \cite{liu2020reactnet}&	1/1&	65.5\\
    Real-to-Bin \cite{martinez2020training}&	1/1&	65.4&	FDA-BNN \cite{xu2021learning}&	1/1&	60.2 \\
    IR-Net \cite{qin2020forward}&	1/1&	58.1&	FDA-BNN* \cite{xu2021learning}&	1/1&	66.0 \\
    SLB \cite{yang2020searching}&	1/1&	61.3&	\textbf{BCDNet-A (R18)}&	\textbf{1/1}&	\textbf{66.9} \\
    SLB \cite{yang2020searching}&	1/8&	66.2&	\textbf{BCDNet-B (R18)}&	\textbf{1/1}&	\textbf{67.9} \\
    \hline
    \end{tabular}
    }
    \end{center}
\end{table}

\subsubsection{Generalization to ResNet-18.} 
The proposed binary MLP block can be easily used as plug-in blocks and improve contextual modeling ability in many BNNs. In Table \ref{Table2}, we compare results in the ResNet-18 (R18) architecture. We replace the last three convolutional layers by 9 binary MLP blocks so as to keep the same operations. Compared with ReActNet (R18), without changing the first three stages, BCDNet-A (R18) improves 1.4\% top-1 accuracy with binary MLPs. For BCDNet-B (R18), significant improvement (+1.0\%) is obtained with the help of dynamic embeddings.

\subsubsection{Generalization to Fine-grained Datasets.} 
As shown in Table \ref{table3}, we also report results on 5 fine-grained datasets: CUB-200-2011 \cite{wah2011caltech}, Oxford-flowers102\cite{nilsback2008automated}, Aircraft\cite{maji2013fine}, Stanford-cars\cite{krause20133d}, Stanford-dogs\cite{khosla2011novel}. Detailed training strategies are reported in Supplementary C. These datasets have much less images than ImageNet but more concentrate on detailed differences of attributions. As shown in Table \ref{table3}, BCDNet-A exceeds ReActNet-A 1.5\% average accuracy, and are able to work well in most cases. We also report results with ImageNet pretraining. Pretraining helps BCDNet-A more than pure binary convolution based ReActNet-A. Modeling contextual dependencies helps BCDNet-A to exceed ReActNet-A by 5.3\% on average, which shows BCDNet is more friendly to pretraining.

\begin{table}[t]
    \caption{Evaluation on fine-grained classification benchmarks. "\dag" indicates the comparison with ImageNet pretraining.}
    \label{table3}
    \begin{center}
    \begin{tabular}{l|c|p{1.4cm}<{\centering}p{1.4cm}<{\centering}p{1.4cm}<{\centering}p{1.4cm}<{\centering}p{1.4cm}<{\centering}|r}
    \hline
    Method&	W/A&	CUB-200&	Flowers&	Aircraft&	Cars&	Dogs&	Avg. \\
    \hline
    ResNet18&	32/32&	52.3&	44.2&	66.8&	39.2&	47.1&	49.9 \\
    BiRealNet18&	1/1&	19.5&	27.0&	18.6&	14.8&	17.9&	19.6 \\
    ReActNet-A&	1/1&	29.9&	29.8&	18.4&	19.4&	18.7&	23.2 \\
    BCDNet-A&	1/1&	34.3&	25.0&	18.1&	25.5&	20.7&	24.7 \\
    \hline
    ReActNet-A\dag&	1/1&	83.7&	80.8&	74.2&	85.4&	70.3&	78.9 \\
    BCDNet-A\dag&	1/1&	90.6&	83.0&	81.1&	91.1&	75.2&	84.2 \\
    \hline
    \end{tabular}

    \end{center}
    \end{table}

\subsection{Ablation Studies}

\begin{minipage}{\textwidth}
\begin{minipage}[t]{0.45\textwidth}
\makeatletter\def\@captype{table}
\begin{tabular}{lcr}
    \hline
    Method & Stages & Top1 Acc. \\
    \hline
    ReActNet-18 &--    & 61.89\%      \\
    BCDNet-18 & 4     & 63.58\%      \\
    BCDNet-18 & 3, 4  & 63.47\%      \\
    BCDNet-18 & 1, 2, 3, 4  & 63.10\%  \\
    \hline
\end{tabular}
\caption{Replacement stage settings}
\label{Table4}
\end{minipage}
\begin{minipage}[t]{0.54\textwidth}
\makeatletter\def\@captype{table}
\begin{tabular}{lcr}
    \hline
    Setting     & Block OPs ${(\times{10}^6)}$    & Top1 Acc. \\
    \hline
    BiConv           & 3.813 ($\times{1.00}$) & 61.89\%  \\
    BiMLP$\times{1}$ & 1.505 ($\times{0.39}$) & 61.32\%  \\
    BiMLP$\times{2}$ & 3.011 ($\times{0.79}$) & 62.64\%  \\
    BiMLP$\times{3}$ & 4.516 ($\times{1.18}$) & 63.58\%  \\
    \hline
\end{tabular}
\caption{Replacement number settings}
\label{Table5}
\end{minipage}
\end{minipage}

\begin{figure}[t]
    \begin{center}
    \includegraphics[width=0.7\linewidth]{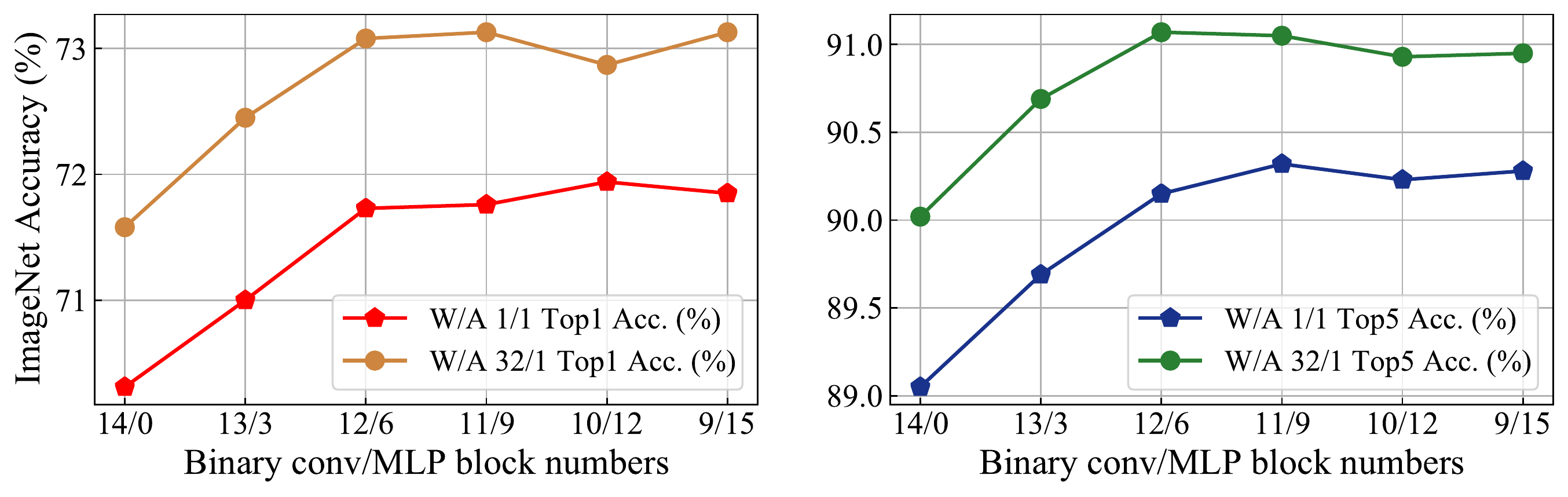}
    \end{center}
    \caption{Performance of replacing binary convolutional blocks with binary MLP blocks. We start from full convolutional ReActNet-A.}
\label{fig:fig5}
\end{figure}

\subsubsection{Efficiency of Binary MLPs.} 
We first study where to replace binary convolutional blocks as proposed binary MLP blocks (BiMLPs) in a ReActNet (RN18), and then report the influence of different configuration of replacements. Training settings in Table \ref{Table4}, \ref{Table5} are reported in Supplementary A. In Table \ref{Table4}, we gradually replace all $3\times3$ binary conv stages of a ReActNet RN18 (except downsampling layers). In practice, we use the same MLP sampling range settings as the last stage MLPs. Results exceed original ReActNet in each drop-in replacement. We find only replacing the last stage is the most efficient. In Table \ref{Table5}, we replace every $3\times3$ binary convolutions with 1, 2 or 3 BiMLPs in the last stage of a ReActNet RN18 (except the downsampling layer). One BiMLP has 0.39 times single block OPs, while drops accuracy slightly; three BiMLPs have 1.18 times block OPs while significantly improve accuracy. In the MobileNet-V1 backbone, we simply replace one convolutional block by three binary MLP blocks once at a time in the ReActNet-A. Fig. \ref{fig:fig5} presents comparisons between MLPs and convolutions, where overall OPs are basically the same ($1.06\times10^8\sim1.08\times10^8$). Due to the influence of randomness, 9$\sim$12 MLP blocks make out the best for a ReActNet-A architecture. For simplicity, we just choose the setting of 11 (1,1,2,2,4,1 blocks for each resolution) convolutional and 9 MLP blocks in other experiments.

\begin{table}[t]
    \caption{Evaluation of different range modeling. We report checkpoint results for different combinations of long- and short-range branches.}
    \label{Table6}
    \begin{center}
    \setlength{\tabcolsep}{2mm}{
    \begin{tabular}{l|cccc}
    \hline
    Traing time&	Stage1&	Stage1+200K&	Stage1+400K&	Stage1+600K \\
    Bitwidth W/A&	32/1&	1/1&	1/1&	1/1 \\
    \hline
    Convolution&	71.58/90.02&	66.30/86.61&	68.27/87.84&	70.31/89.05 \\
    P-L-L&	72.85/90.82&	67.06/87.32&	69.14/88.42&	70.79/89.39 \\
    P-S-S&	71.98/90.56&	68.24/88.10&	69.87/89.07&	71.72/89.97 \\
    P-S-L&	\textbf{73.13/91.05}&	\textbf{68.63/88.45}&	\textbf{70.46/89.37}&	\textbf{71.76/90.32} \\
    \hline
    \end{tabular}
    }
\end{center}
\end{table}

\subsubsection{Evaluation of Contextual Dependencies.} 
Contextual dependencies come from long-short range sampling and the MLP architecture. In Table \ref{Table6}, we keep three branches in every binary MLP block and evaluate different sampling range combinations. ‘P’, ‘S’ and ‘L’ indicate the pointwise (original location) branch, the short-range branch and the long-range branch, with token sampling ranges of 0, 1 and half of the resolution respectively. We also report results of 200k, 400k and 600k training iterations. In Table \ref{Table6}, combination of long-short ranges (“P-S-L”) with different receptive field achieves better performance. Compared with only long ranges (“P-L-L”), “P-S-L” has inductive bias and achieve better performance. Compared with only short ranges (“P-S-S”), “P-S-L” places long-range modeling and can be easier to converge in early training times (e.g., 200k, 400k iterations). In the MLP architecture, even though without long-ranges, “P-S-S” still exceeds binary CNNs because it decouples a $3\times3$ binary convolution to three BiMLPs and expands reception field in each layer (from $3\times3$ to $7\times7$).

\begin{figure}[t]

    \begin{center}
    \includegraphics[width=0.7\linewidth]{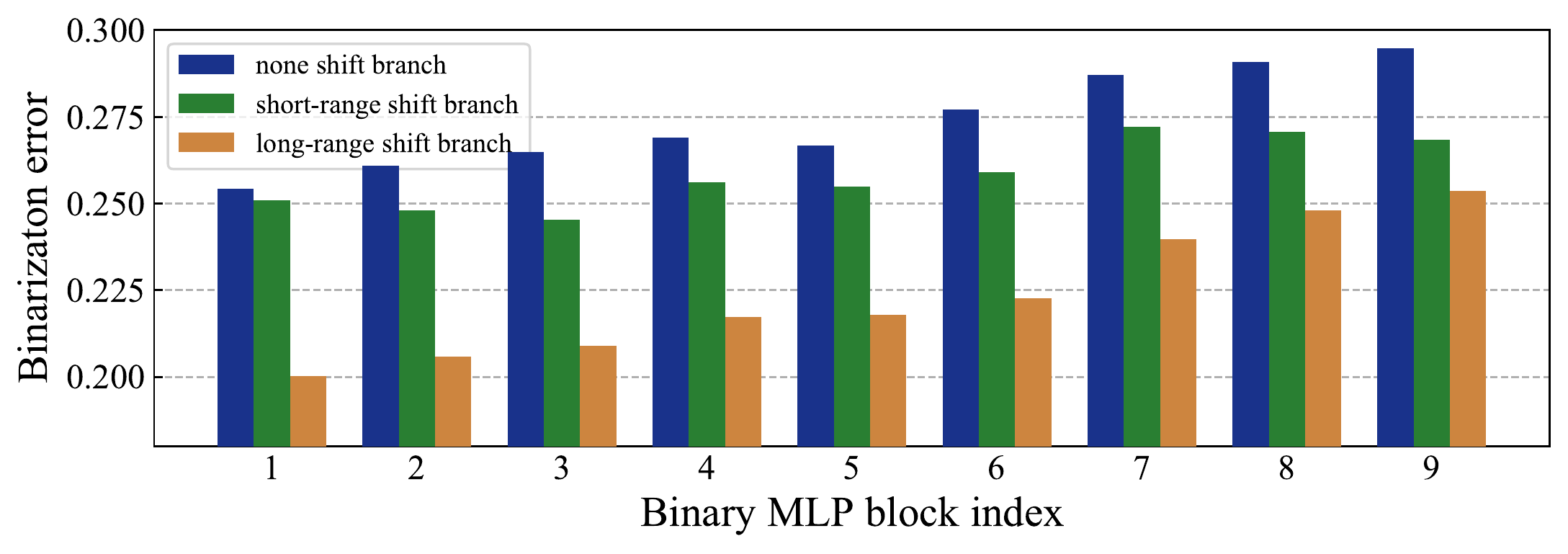}
    \end{center}
    \caption{Binarization error of different branches in MLP layers. 
    }
    \label{fig:fig4}
    \end{figure}

\subsubsection{Binarization Error Analysis.}
We define the average binarization error as:
\begin{eqnarray}  \label{eq17}
    error=\frac{1}{n}\sum | \frac{{\left \| sign(\mW) \right \|}_{\ell_1}}{c_{in}} sign(\mW)-\mW| 
\end{eqnarray}
where $\frac{{\left \| sign(\mW) \right \|}_{\ell_1}}{c_{in}}$ follows Eq. \ref{eq17}; $\mW$ is the weight matrix in a binary MLP layer with $n$ numbers. For BCDNet-A, we calculate the average binarization error of each MLP layer in Fig. \ref{fig:fig4}. Note that, when binarization error increases, gradient estimation is inaccurate when backpropagation. We find in almost every block, errors in 3 branches follows a order of: $\texttt{pointwise} \textgreater \texttt{short-range} \textgreater \texttt{long-range}$. Binarization error is improved by decreasing of exchange distance, which indicates the long-range branch is easier to optimize than the short-range branch.

\section{Conclusion}
This work improves binary neural networks by modeling contextual dependencies. A long-short range binary MLP module is proposed to explore comprehensive contextual perceptions, which can be an efficient alternative to traditional binary convolutions. Equipped with our binary MLPs and improved binary convolutions, BCDNets exceed state of the arts significantly.

\section*{Acknowledgements}
This work was supported by the National Natural Science Foundation of China (U20B2042), the National Natural Science Foundation of China (62076019) and Aeronautical Science Fund (ASF) of China (2020Z071051001).

\clearpage
%
%

\end{document}